\documentclass[conference]{IEEEtran}
\IEEEoverridecommandlockouts

\usepackage{cite}
\usepackage{amsmath,amssymb,amsfonts,bm}
\usepackage{algorithmic}
\usepackage{graphicx}
\usepackage{textcomp}
\usepackage{xcolor}
\usepackage{soul} 
\usepackage{afterpage}         
\usepackage{dblfloatfix}
\sethlcolor{yellow}

\usepackage{xspace}
\newcommand{\robotname}{\textsc{Manipulider}\xspace}

\def\BibTeX{{\rm B\kern-.05em{\sc i\kern-.025em b}\kern-.08em
    T\kern-.1667em\lower.7ex\hbox{E}\kern-.125emX}}

\begin{document}

\title{Manipulider: A Multi-Engine Buoyancy-Controlled Robot for Thrusterless Underwater Gliding and Manipulation}


\author{%
Yitao Jiang\textsuperscript{1},
Yewei Huang\textsuperscript{1},
Weizhi Cao\textsuperscript{2},
Mingi Jeong\textsuperscript{3},
Alberto Quattrini Li\textsuperscript{1},
Luyang Zhao\textsuperscript{4},\\
Muhao Chen\textsuperscript{2},
Devin Balkcom\textsuperscript{1}%
}

\maketitle
\begingroup
\renewcommand\thefootnote{\arabic{footnote}}
\setcounter{footnote}{0}
\footnotetext[1]{Dept. of Computer Science, Dartmouth College, Hanover, NH, USA. Email: \{yitao.jiang.gr, yewei.huang, alberto.quattrini.li, devin.balkcom\}@dartmouth.edu}
\footnotetext[2]{Dept. of Mechanical and Aerospace Engineering, University of Houston, Houston, TX, USA. Email: \{wcao15, muhaochen\}@uh.edu}
\footnotetext[3]{Dept. of Aerospace and Ocean Engineering, Virginia Tech, Blacksburg, VA, USA. Email: mingijeong@vt.edu}
\footnotetext[4]{Dept. of Electrical and Computer Engineering, Clemson University, Clemson, SC, USA. Email: luyangz@clemson.edu}
\endgroup

\begin{abstract}
The \robotname is a buoyancy-actuated underwater robot that enables thrusterless, glide-like locomotion and attitude-based manipulation, while providing a magnetic modular interface for rapid payload swapping (e.g., a gripper or sensors). Four syringe-based buoyancy engines distributed around the body jointly regulate net buoyancy and the center of buoyancy, allowing the vehicle to maintain large tilt angles through static force balance without continuous thrust and to avoid propeller entanglement risks. We present the mechanical and electrical design, calibration procedure, and control architecture. Experiments with a gripper attached (no external payload) show a controllable buoyancy-displacement range of 40~mL per engine (\(\approx\)160~g total buoyancy authority), maximum statically stable tilts of \(64.6^{\circ}\) (single-engine) and \(61.8^{\circ}\) (dual-engine), and representative vertical and tilt-transition dynamics. We further demonstrate tilt regulation, controlled ascent/descent primitives, and a proof-of-concept gripper-based payload-transport sequence without thrusters.
\end{abstract}


\section{Introduction}
\label{sec:introduction}

Underwater robots are increasingly deployed for inspection, monitoring, and intervention tasks in complex natural environments. However, long-duration operation and reliable mobility remain challenging due to energy constraints and the difficulty of maintaining stable posture in the presence of currents and obstacles \cite{aracri_soft_2021,li_bioinspired_2023,qu_recent_2023}.

Most autonomous underwater vehicles (AUVs) use thrusters (and sometimes control surfaces, such as fins) to regulate depth and attitude. Thruster-driven platforms can be highly maneuverable, but are inherently power-inefficient and can produce significant disturbance and noise \cite{macauley_reefglider_2024}, because they need continuous actuation to maintain a particular attitude. In cluttered environments, propellers may also be susceptible to interference or entanglement, which can compromise reliability \cite{irgens2021experimental}. These limitations motivate alternative actuation strategies that can maintain posture with low steady-state power.

Buoyancy-driven systems offer an appealing direction because they can change depth by regulating displaced volume, enabling efficient long-range motion in the style of underwater gliders \cite{mitchell_developing_2013,williams_design_2018}. Recent work has also explored compact and soft implementations, including fluidic closed-loop control for untethered gliders \cite{bonofiglio_soft_2023}. Nevertheless, many buoyancy-driven robots achieve limited volume change relative to vehicle size and therefore have limited buoyancy authority, which constrains payload capacity and the range of achievable operating conditions. In addition, attitude control is often realized by shifting the center of mass (e.g., translating an internal battery), which can provide only a limited posture range and controllability~\cite{leonard2001model,graver2005underwater,petritoli2024drones,WU20211099}. As a result, many platforms still rely on thrusters or auxiliary mechanisms to achieve agile posture changes or to robustly manage ascent and descent.

\begin{figure}[t]
  \centering
  \includegraphics[width=\columnwidth]{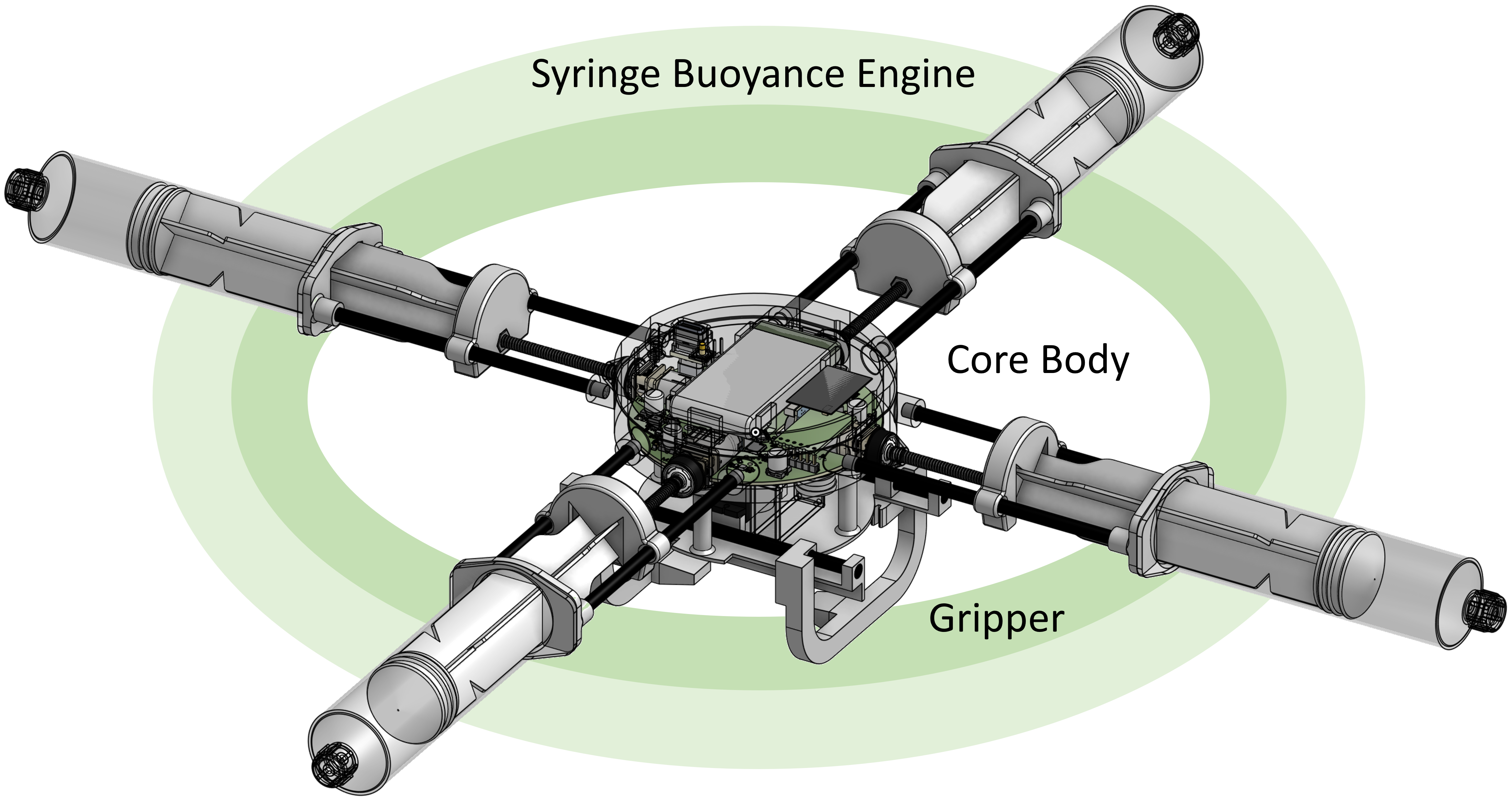}
  \caption{System overview of \robotname (CAD rendering). The robot consists of a sealed central body with four syringe-based buoyancy engines arranged symmetrically around the frame, and a magnetically attached gripper module mounted underneath. The multi-engine layout enables both net buoyancy modulation and differential buoyancy distribution for posture control.}
  \label{fig:system_overview}
\end{figure}

This paper presents \robotname (Fig.~\ref{fig:system_overview}), a buoyancy-actuated underwater robot that combines thrusterless, glide-like locomotion with attitude-based payload manipulation and modular payload integration. Four buoyancy engines distributed around the body jointly regulate net buoyancy and the buoyancy center, enabling statically stable, whole-body tilting without continuous thrust. A magnetic payload interface mounted underneath supports rapid swapping of task-specific modules, such as a gripper or sensors.

At the mechanism level, \robotname adopts a multi-engine variable-buoyancy architecture based on syringe-type volume change. Compared with a single large variable-volume chamber, distributing buoyancy actuation across multiple smaller engines reduces the pressure-loaded area per actuator, thereby lowering the required actuation force at depth and relaxing material strength requirements. Moreover, buoyancy margin and payload capability can be increased by adding engines, offering a scalable design path.

The main contributions of this work are:
\begin{itemize}
  \item A multi-engine buoyancy-controlled underwater robot that jointly regulates net buoyancy and the buoyancy center for wide-range, statically stable whole-body tilting.
  \item A control architecture for stable hovering at multiple attitudes and thrusterless, glide-like translation primitives via coordinated buoyancy and posture commands.
  \item A magnetic modular payload interface enabling rapid integration of task-specific modules, with gripper-based manipulation demonstrations.
  \item An experimental characterization of the robot’s performance, quantifying a usable buoyancy displacement of 10--50~mL per engine, corresponding to a 40 mL stroke per engine and \(\approx\)160~g total buoyancy authority, maximum statically stable tilts of 64.6$^\circ$ (single-engine) and 61.8$^\circ$ (dual-engine), and representative vertical/tilt transition dynamics.
\end{itemize}

\begin{table}[h]
\caption{Summary of the \robotname prototype characteristics.}
\label{tab:performance_summary}
\centering
\begin{tabular}{l c}
\hline
Metric & Value \\
\hline
Base robot mass & 399.2~g \\
Gripper module mass & 86.7~g \\
\hline
Usable displacement range per engine & 10--50~mL \\
Total buoyancy authority & \(\approx\)160~g \\
\hline
Max statically stable tilt angle (single-engine) & 64.6$^\circ$ \\
Max statically stable tilt angle (dual-engine, diagonal) & 61.8$^\circ$ \\
\hline
Representative tilted-hover equilibrium tilt angle & 0.921~rad \\
Peak tilt rate during transition & 0.332~rad/s \\
Peak tilt acceleration during transition & 0.528~rad/s$^2$ \\
\hline
Ascent max speed & 56.2~mm/s \\
Ascent max acceleration & 362.3~mm/s$^2$ \\
Descent max speed & 95.5~mm/s \\
Descent max acceleration & 370.0~mm/s$^2$ \\
\hline
\end{tabular}
\end{table}

\section{Related Work}
\label{sec:related_work}

This section reviews prior work on thruster-driven AUVs and buoyancy-driven gliders, variable-buoyancy mechanisms and balance control, attitude control via internal mass shifting versus distributed buoyancy, and modular payload interfaces for underwater manipulation. We emphasize how \robotname differs in actuation design and control authority, particularly for multi-attitude hovering without continuous thrust.

\subsection{Thruster-Driven AUVs and Buoyancy-Driven Gliders}
Thruster-driven AUVs provide agile maneuverability and are well-suited for operation in complex environments; however, they typically incur substantial energy consumption and can introduce non-negligible hydrodynamic disturbances. In contrast, underwater gliders achieve efficient long-range motion by cycling buoyancy and commonly rely on internal mass shifting and/or control surfaces for attitude regulation \cite{mitchell_developing_2013,williams_design_2018}. Classic glider systems and models established the canonical framework in which
periodic changes in buoyancy and internal mass distribution convert vertical excursions into horizontal displacement with ultra-low power consumption \cite{eriksen2001seaglider,sherman2001spray,webb2001slocum,rudnick2004mtsj,leonard2001model,graver2005underwater}.
Follow-up work improved depth regulation by increasing the accuracy and repeatability of buoyancy actuation and associated sensing
\cite{petritoli2019sensors}. These foundations underscore the endurance advantage of buoyancy-driven locomotion, while also highlighting limitations in maneuverability and fine-grained posture authority, especially for small-scale platforms. Hybrid systems have also combined buoyancy engines with propulsive actuation for depth control~\cite{snyder2018hybrid}.

Recent efforts aim to narrow this gap by improving maneuverability while retaining the efficiency benefits of buoyancy actuation. For example, ReefGlider proposes a platform based on a vectored buoyancy engine, targeting enhanced maneuverability under buoyancy-only control \cite{macauley_reefglider_2024}. \robotname shares the motivation to reduce reliance on thrusters, but focuses on distributed buoyancy engines that directly modulate the center of buoyancy to enable wide-range hovering and controlled glide transitions under a
common actuation suite, while also emphasizing modular payload integration.

\subsection{Variable Buoyancy Mechanisms and Buoyancy/Balance Control}
Buoyancy control has been explored through bio-inspired mechanisms, such as the phase change of spermaceti oil in sperm whales \cite{shibuya_underwater_2006}. Other designs develop compact variable-buoyancy devices for long-duration vertical profiling and low-power locomotion, including soft and fluidic actuation and closed-loop control \cite{bonofiglio_soft_2023,fornai_design_2013}.
Research has also addressed buoyancy and balance regulation during dynamic payload changes. Detweiler \emph{et al.} present mechanisms and control algorithms that adapt buoyancy and balance without increasing thruster workload \cite{siciliano_saving_2009}. Related hovering platforms explore buoyancy and balance control to reduce steady thruster usage under changing payload conditions (e.g., AMOUR~V) \cite{vasilescu2010amour}.

Beyond single-vehicle demonstrations, recent work increasingly frames variable-buoyancy systems (VBS) as reusable subsystems for small AUVs and gliders. Electromechanical and electrohydraulic buoyancy-change modules have been developed and experimentally evaluated, exploring trade-offs among authority, packaging, and robustness \cite{carneiro2022actuators,carneiro2023actuators}. A comparative study further quantified the dynamic response and power requirements of VBS versus propeller-based actuation, reinforcing the efficiency advantage of buoyancy-driven depth control for long-duration missions \cite{carneiro2024sensors}. Additional studies examine VBS design and
control for hovering and miniature implementations, informing practical design choices at the small scale \cite{tiwari2020jmse,elkolali2022thesis}. While these efforts improve repeatable buoyancy actuation and depth regulation, many
modules primarily target endurance and profiling rather than wide-range, statically stable attitude control via buoyancy distribution. Thermal energy engines offer an orthogonal route to endurance \cite{yang2016applthermaleng}, but do not directly address agile attitude authority through rapid and precise buoyancy redistribution.

\subsection{Attitude Control: Mass Shifting vs.\ Distributed Buoyancy}
A common approach for buoyancy-driven vehicles is to shift internal masses to change the center of mass and thereby regulate pitch and roll, often in combination with buoyancy cycling and control surfaces \cite{eriksen2001seaglider,sherman2001spray,webb2001slocum,leonard2001model,graver2005underwater}. Mass-based reorientation has also been used in profiling floats, such as Flippin'\(\chi\)SOLO~\cite{moum2023flippin}.
While effective for many glider missions, the achievable attitude range and control bandwidth are often constrained by mechanical travel, internal packaging, and the accuracy/repeatability limits of buoyancy and depth regulation \cite{graver2005underwater,petritoli2019sensors}. In contrast, distributed buoyancy actuation directly changes the center of buoyancy (CoB), enabling buoyancy-generated restoring torques that can maintain nontrivial attitudes through static force balance. ReefGlider provides an example of enhancing control authority via buoyancy-based actuation elements \cite{macauley_reefglider_2024}. \robotname contributes a compact, syringe-based multi-engine design and targets a wide range, statically stable multi-attitude hovering without continuous thrust by coordinating both net
buoyancy and CoB.

\subsection{Modularity, Magnetic Interfaces, and Underwater Manipulation}
Modularity and magnetic coupling have been used in underwater robotics to improve reconfigurability while mitigating sealing challenges. For example, MMBAUV uses magnetically coupled modules to enable modular bio-inspired swimming while addressing watertightness constraints \cite{wright_design_2023}. Magnetic alignment has also been explored for underwater docking to facilitate recharging or reconfiguration \cite{mintchev_towards_2015}. \robotname complements this direction by providing a magnetic payload interface intended for rapid swapping of task-specific modules beneath a common buoyancy-and-control core.

In parallel, underwater manipulation has been explored using mobile manipulation platforms and compliant arms/grippers
\cite{liu_underwater_2020,wu_glowing_2022}. \robotname is positioned as a buoyancy-actuated platform that can potentially support such tasks by enabling multi-attitude hovering and a repeatable, rapid payload mounting interface, while maintaining thrusterless operation during hovering and gliding.

\subsection{Summary of Differences}
In summary, prior work establishes buoyancy-driven locomotion as an energy-efficient alternative to thruster-centric AUVs and explores a range of variable-buoyancy mechanisms and balance control strategies
\cite{mitchell_developing_2013,williams_design_2018,bonofiglio_soft_2023,siciliano_saving_2009}.
\robotname differs by combining (i) distributed buoyancy engines for direct center-of-buoyancy control, (ii) wide-range, statically stable multi-attitude hovering without continuous thrust, (iii) controlled hover-to-glide behaviors through coordinated buoyancy and posture, and (iv) a magnetic modular payload interface aimed at rapid task adaptation.

\section{Methodology}
\label{sec:methodology}

\subsection{System Overview}
\label{sec:method_overview}
Figure~\ref{fig:system_overview} gives a system-level overview of \robotname. The robot consists of a sealed central core body and four identical syringe buoyancy engines arranged symmetrically around the body. Each buoyancy engine is driven by an independent motor, enabling both net buoyancy control and differential buoyancy distribution for posture regulation. A magnetic snap interface on the underside of the body enables rapid attachment of task-specific modules (e.g., a gripper or sensors), allowing \robotname to serve as a reconfigurable underwater platform. The magnetic connector also provides electrical contacts for supplying power and communication to the attached payload module, enabling rapid functional reconfiguration without modifying the sealed core.

The CAD model (Fig.~\ref{fig:system_overview}) has an overall footprint of approximately 565~mm \(\times\) 565~mm; the central body is 26~mm thick, approximately 33~mm including the top sealing port, and the gripper is 63~mm tall with a maximum grasp width of 61~mm.

\subsection{Buoyancy Control Concept}
\label{sec:method_concept}
\begin{figure}[t]
  \centering
  \includegraphics[width=\columnwidth]{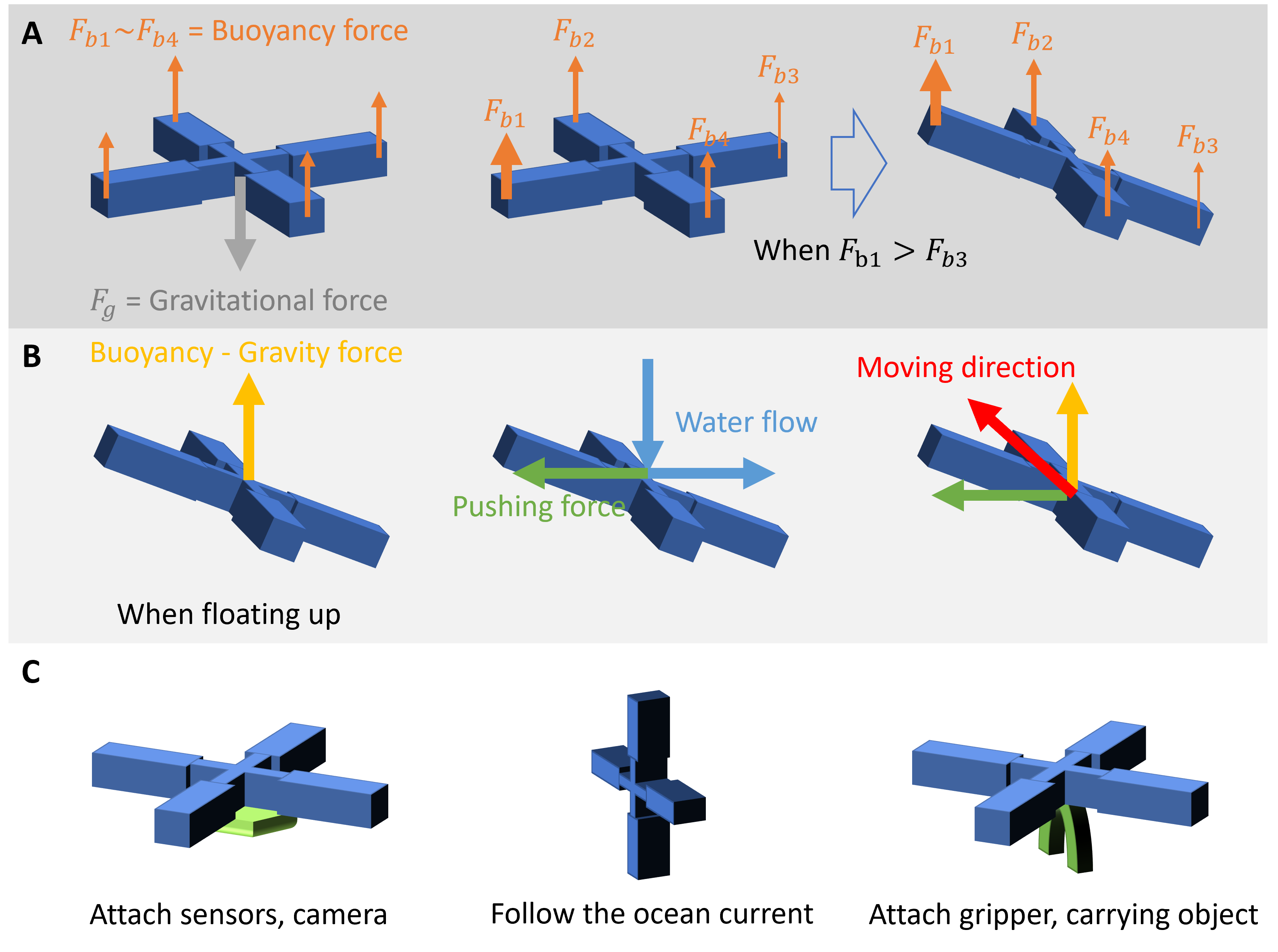}
  \caption{Concept illustration of distributed buoyancy control. (A) Differential buoyancy across the four engines shifts the center of buoyancy and generates attitude torques. (B) Coordinating net buoyancy with a pitched attitude enables thrusterless gliding. (C) Magnetic attachment enables rapid payload changes, supporting different sensing or manipulation modules in concept.}
  \label{fig:concept}
\end{figure}

We control \robotname's motion through the buoyancy forces produced by the four engines. As illustrated in Fig.~\ref{fig:concept}(A), changing the relative buoyancy across the four arms shifts the center of buoyancy and produces a restoring torque that sets the robot's attitude. Unlike thruster-based attitude control, which typically requires continuous power and is vulnerable to propeller entanglement in cluttered environments \cite{irgens2021experimental}, the proposed approach can maintain a desired attitude through static force balance with low steady-state energy consumption.

Net buoyancy controls vertical motion. When the total buoyancy exceeds the weight, the robot ascends; when it is lower, the robot descends. As shown in Fig.~\ref{fig:concept}(B), if the robot holds a nonzero pitch angle while ascending or descending, hydrodynamic forces generate a horizontal component of motion, enabling thrusterless gliding. This mode can be used for energy-efficient translation without propellers. Figure~\ref{fig:concept}(C) highlights the modular payload concept: by attaching different payloads beneath the body via magnets, \robotname can be adapted for a range of tasks. When equipped with a gripper, the platform can support proof-of-concept object transport and future manipulation primitives, which we discuss further in Sec.~\ref{sec:limitations}.

\subsection{Mechanical Design and Assembly}
\label{sec:method_mechanical}

\begin{figure}[t]
  \centering
  \includegraphics[width=\columnwidth]{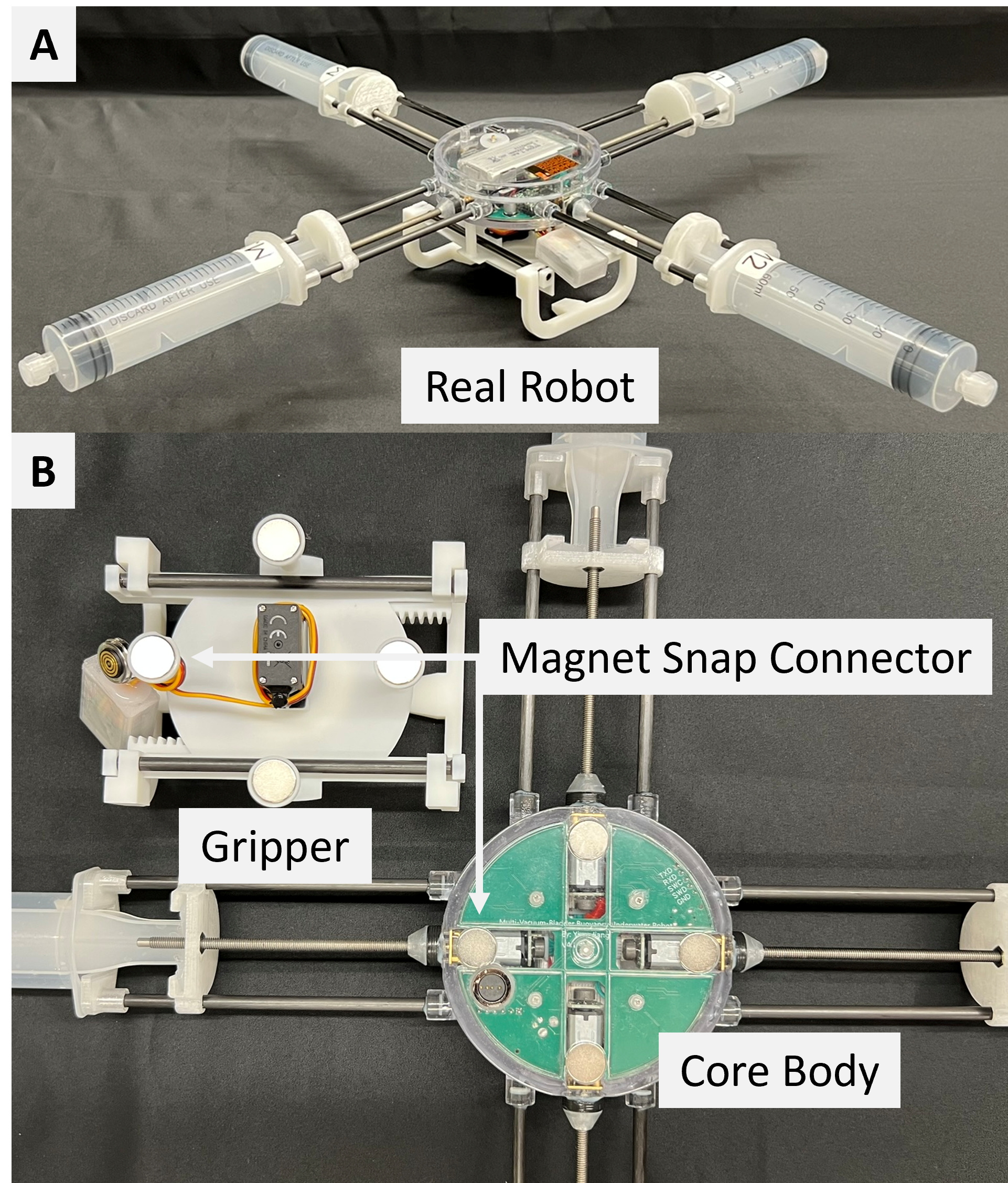}
  \caption{\robotname prototype and magnetic snap interface. (A) Assembled robot. (B) Underside view showing the magnetic snap connector, example gripper module, and the core body.}
  \label{fig:real_robot}
\end{figure}

\begin{figure}[t]
  \centering
  \includegraphics[width=\columnwidth]{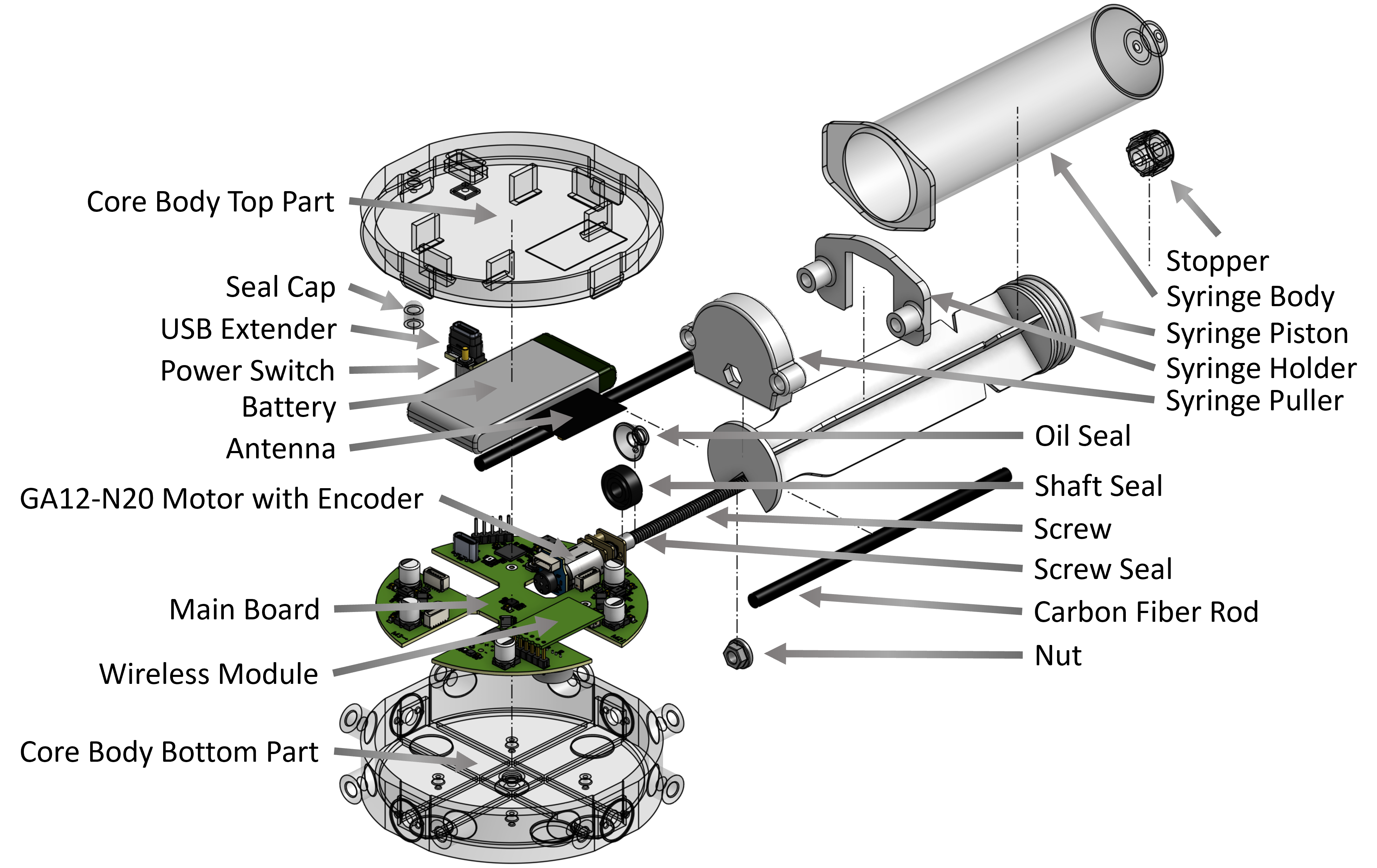}
  \caption{Exploded assembly view of the core body and one syringe buoyancy engine, highlighting the actuation transmission and sealing components used to maintain water tightness.}
  \label{fig:assembly}
\end{figure}

The physical robot is shown in Fig.~\ref{fig:real_robot}. The central core body houses the battery, controller, motor drivers, and sensors in a sealed enclosure. The magnetic snap connector on the underside uses a commercial 4-pin, 360$^\circ$ self-aligning waterproof magnetic connector, enabling rapid attachment and detachment of payload modules while providing repeatable alignment and electrical contacts. The connector was used in shallow pool experiments; pressure-rated or long-duration saltwater deployment remains a future engineering refinement.

Figure~\ref{fig:assembly} shows an exploded view of the assembly. Each buoyancy engine is implemented using a sealed 60~mL syringe whose plunger is actuated by a motor-driven lead screw. The designed operational range is 10--50~mL; the lower margin supports trimming and neutral-buoyancy calibration, while the remaining stroke provides a mechanical safety margin. Guide rods constrain the moving elements, reducing side loading and preventing binding during stroke. Multiple seals are used along the load path to preserve water tightness while permitting motion.

We further characterize the volumetric actuation rate of each syringe engine, \(\dot{V}\), which directly affects the buoyancy change rate \(\dot{F}_b = \rho g \dot{V}\). A representative 10--30~mL sweep within the operational range yielded average volume change rates of 0.98843~mL/s and 1.04801~mL/s in the two directions, respectively. This sweep was used for rate characterization and calibration, not to define the full 10--50~mL operating range.

A key design choice is to use four identical engines rather than a single large variable-volume chamber. Distributing the pressure-loaded area across multiple smaller syringes reduces the peak force demanded from any single actuator at depth and lowers structural requirements. It also enables differential buoyancy commands for attitude control, rather than relying solely on internal mass shifting\cite{leonard2001model, petritoli2019sensors}.

The internal components are arranged as symmetrically as possible about the body center, and minor trimming mass is used during calibration to make the center of mass close to the geometric center of the vehicle. This reduces constant gravitational bias torques and improves the repeatability of buoyancy-based attitude control.

\subsection{Electronics and Control Architecture}
\label{sec:method_electronics}

\begin{figure}[t]
  \centering
  \includegraphics[width=\columnwidth]{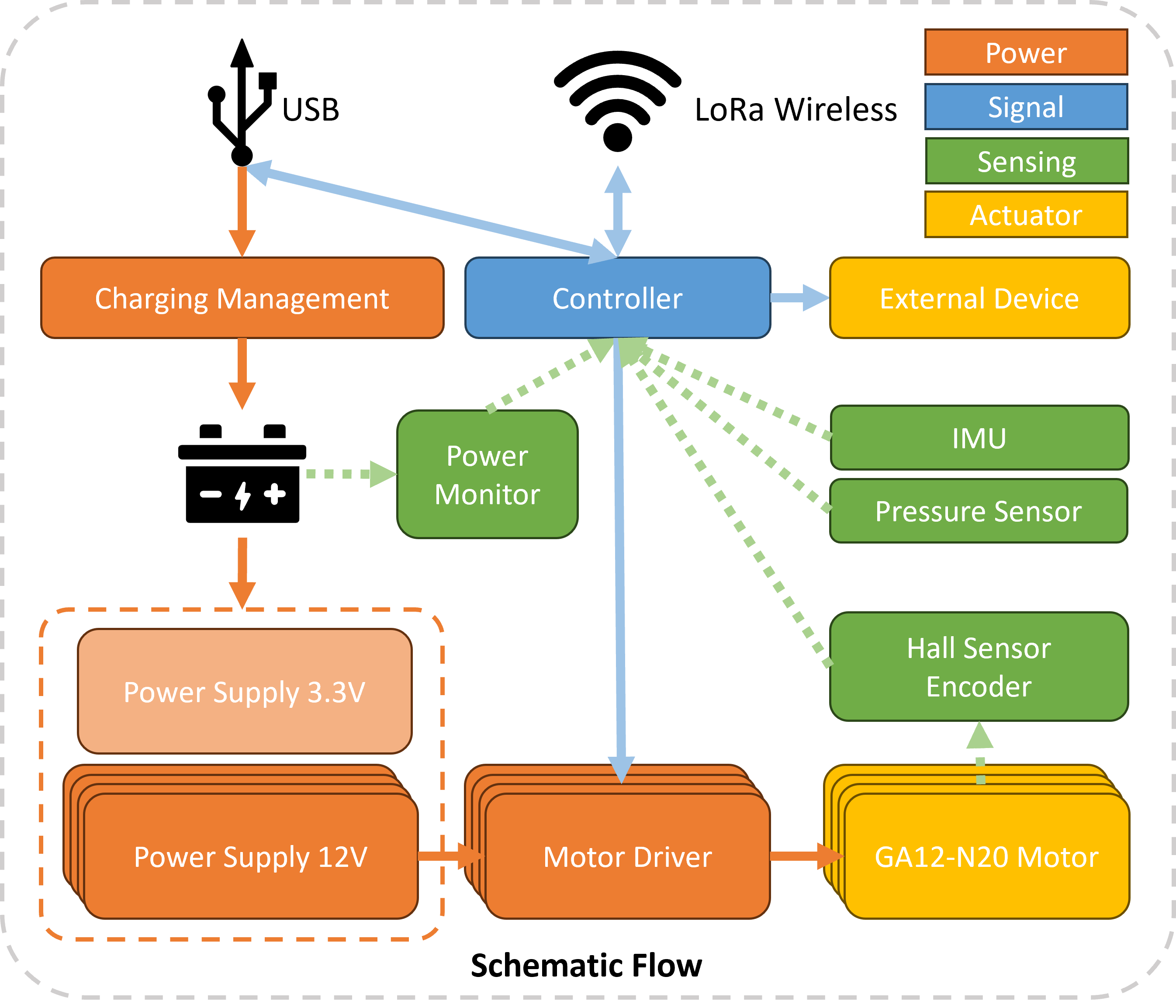}
  \caption{Electronics architecture of \robotname. The controller reads IMU and pressure sensors for attitude and depth estimation, monitors power, and commands four motor drivers (one per buoyancy engine) using encoder feedback for closed-loop position control.}
  \label{fig:architecture}
\end{figure}

The electronics architecture is summarized in Fig.~\ref{fig:architecture}. The system is built around an RP2040 microcontroller, a 6-axis IMU, an MS5837 pressure sensor, and four GA12-N20 brushed DC motors with rear magnetic encoders and 100~mm lead-screw output shafts. The robot is powered by an 803060 lithium-ion polymer battery with a nominal capacity of 2000~mAh and a 25C discharge rating. Power is split into four boosted motor rails at approximately 11.5~V (nominally 12~V), one for each syringe motor, and a regulated 3.3~V rail for the controller, sensors, wireless module, and logic electronics. System voltage and current are monitored onboard, and a wireless UART link supports command and telemetry exchange. Each buoyancy engine is driven by a dedicated motor driver using encoder feedback, enabling closed-loop control of plunger position and repeatable displaced-volume changes. On the RP2040, sensing/communication and motor control are split across the two cores: Core~0 periodically acquires sensor data and services the wireless command/status link, while Core~1 runs a dedicated 1~kHz motion-control loop for the four syringe actuators. This separation prevents communication and sensor I/O latency from interfering with actuator position control.

\subsection{Dynamics and Attitude Control Formulation}
\label{sec:method_control}
We define the robot pose $\bm{\eta}$ and body-fixed velocity $\bm{\nu}$ as
\begin{align}
\bm{\eta} &=
\begin{bmatrix}
\bm{p} \\
\bm{\Theta}
\end{bmatrix}
=
\begin{bmatrix}
x & y & z & \phi & \theta & \psi
\end{bmatrix}^T, \\
\bm{\nu} &=
\begin{bmatrix}
\bm{v} \\
\bm{\omega}
\end{bmatrix}
=
\begin{bmatrix}
u & v & w & p & q & r
\end{bmatrix}^T,
\end{align}
where $\bm{p}$ is the position in the inertial frame, $\bm{\Theta}$ are roll–pitch–yaw angles, and $\bm{\nu}$ is expressed in the body-fixed frame.
The rigid-body dynamics of the robot expressed in the body-fixed frame are
\begin{equation}
M \dot{\bm{\nu}} + \bm{C}(\bm{\nu})\bm{\nu} = \bm{\tau},
\end{equation}
where
{
\footnotesize
$M =
\begin{bmatrix}
mI_{3\times3} & 0 \\
0 & I
\end{bmatrix},$
}
$m$ is the total mass, $I_{3\times3}$ is the $3\times3$ identity matrix, and $I$ is the inertia tensor expressed in the body frame. 
The matrix $\bm{C}(\bm{\nu})$ denotes the rigid-body Coriolis and centripetal matrix associated with the velocity vector $\bm{\nu}$.

The total external wrench in the inertial frame is denoted by $\bm{\tau}^w$.
The corresponding wrench expressed in the body-fixed frame is denoted by $\bm{\tau}$
\begin{align}
\bm{\tau} =
\begin{bmatrix}
R^T & 0 \\
0 & R^T
\end{bmatrix}
\bm{\tau}^w, \ 
\bm{\tau}^w =
\begin{bmatrix}
\bm{F}_{B}^w + \bm{F}_D^w \\
\bm{\tau}_{B}^w + \bm{\tau}_D^w
\end{bmatrix},
\end{align}
where $R \in SO(3)$ is the rotation matrix from body to inertial frame. 
$\bm{F}_{B}^w$ and $\bm{\tau}_{B}^w$ denote the buoyancy force and torque expressed in the inertial frame, 
while $\bm{F}_{D}^w$ and $\bm{\tau}_{D}^w$ denote the drag force and torque expressed in the inertial frame.
We first define the buoyancy force and the corresponding torque generated by the four arms:
\begin{align}
\bm{F}_{B}^w = (F_b - F_g)\bm{e}_z, \ 
\bm{\tau}_{B}^w
=
\sum_{i=1}^4
(R\bm{r}_i) \times (F_{b,i}\bm{e}_z),
\end{align}
where $\bm{e}_z$ denotes the unit vector along the $z$-axis, and $g$ denotes the gravitational acceleration.
Let $F_g$ be the gravitational force (weight) of the robot. 
$\bm{r}_i$ denotes the position vector of the $i$-th arm expressed in the body frame.
We denote the buoyant force produced by engine $i\in\{1,2,3,4\}$ as $F_{b,i}$. The net buoyant force is:
\begin{equation}
F_b = \sum_{i=1}^4 F_{b,i}.
\end{equation}
The sign of $\bm{F}_{B}^w$ determines the vertical motion regime:
\begin{itemize}
  \item \textbf{Neutral buoyancy:} $F_b = F_g$.
  \item \textbf{Floating up:} $F_b > F_g$.
  \item \textbf{Sinking down:} $F_b < F_g$.
\end{itemize}
Since buoyancy is generated by displaced volume, a convenient intermediate variable is the displaced volume change $\Delta V_i$ of each engine, where
\begin{equation}
F_{b,i} = \rho g \Delta V_i,
\end{equation}
with $\rho$ the water density and $g$ gravitational acceleration. In practice, $\Delta V_i$ is controlled by commanding the syringe plunger displacement, which is tracked using motor encoder feedback (Fig.~\ref{fig:architecture}). A representative 10--30~mL sweep yields an approximately linear encoder-to-volume relationship: a 1~mL volume change corresponds to approximately 8138 encoder counts on average, i.e., \(\Delta n \approx 8138\,\Delta V\) with \(\Delta V\) in mL. We use this relationship to convert desired displaced-volume changes into motor position setpoints.

We then define the drag force and the corresponding torque acting on the robot:
\begin{align}
\bm{F}_D^w = \sum_{i=1}^4 \bm{F}_{D,i}, \
\bm{\tau}_D^w
=
\sum_{i=1}^4
(R\bm{r}_i)
\times
\bm{F}_{D,i}
\end{align}
We denote the drag force on each arm as $\bm{F}_{D,i}$. $\bm{F}_{D,i}$ is modeled based on the component of the relative velocity perpendicular $\bm{v}_{\perp,i}$ to $\bm{r}_i$:
\begin{align}
\bm{F}_{D,i}
=
\frac{1}{2}
\rho C_D A
\|\bm{v}_{\perp,i}\|
\bm{v}_{\perp,i}
\end{align}
where $\rho$ is the fluid density and $C_D$ is the dimensionless drag coefficient.
Define the unit vector along $\bm{r}_i$ as $\bm{a}_i = \frac{\bm{r}_i}{\|\bm{r}_i\|}, \bm{a}_i^w = R \bm{a}_i$. Here we assume a uniform ambient current, represented by a constant velocity vector $\bm{v}_c$ in the inertial frame.
The relative velocity at the $i$-th arm in the inertial frame is
\begin{align}
\bm{v}_{rel,i}^w
&=
\bm{v}_c - R \bm{v}_i, \\
\bm{v}_i
&=
\bm{v} + \bm{\omega} \times \bm{r}_i.
\end{align}
The perpendicular component of the relative velocity is given by
\begin{align}
\bm{v}_{\perp,i}
=
\bm{v}_{rel,i}^w
-
(\bm{v}_{rel,i}^w \cdot \bm{a}_i^w)\bm{a}_i^w.
\end{align}

The formulation above captures the dominant buoyancy forces/torques and arm drag used for offline analysis of tilted ascent/descent. In this paper, ``glide-like'' refers to the horizontal motion generated when a tilted body ascends or descends through the surrounding water. To maintain high-frequency control cycles on the resource-constrained embedded controller, these calculations are not integrated into the real-time onboard loop; instead, high-level setpoints are tuned empirically as described below:
\begin{itemize}
  \item A high-level depth controller (external) regulates $F_b$ (common-mode command across the four engines) using pressure feedback.
  \item A high-level attitude controller (external) regulates vehicle tilt using IMU feedback through differential volume commands.
  \item Four low-level motor position loops track individual plunger position commands using encoder feedback.
\end{itemize}

In the current prototype, the onboard firmware provides reliable low-level execution of per-engine syringe displacement commands using motor encoder feedback, enabling repeatable volume changes and static holding. High-level setpoints (net buoyancy changes and differential commands for a desired tilt direction) are tuned empirically during experiments and may vary with payload properties and environmental conditions (e.g., trapped air, water currents, and hydrodynamic interaction). Developing a general, model-based mapping from desired depth/tilt to per-engine setpoints that remains robust across payloads and environments is an important direction for future work (Sec.~\ref{sec:limitations}).

\subsection{Calibration}
\label{sec:method_calibration}
Our current calibration procedure focuses on buoyancy trimming for stable hovering. With all four engines set to a nominal mid-stroke position, we adjust the total mass/buoyancy so that the robot reaches neutral buoyancy in still water and remains suspended. We then fine-tune the trim so that, at rest, the robot hovers with its main body approximately parallel to the ground (i.e., minimal static tilt bias).

\section{Results}

Tilt angles are computed from the onboard IMU accelerometer, while speeds/accelerations are estimated from external video using a calibrated scale reference.

\subsection{Buoyancy Performance}
In representative vertical motion tests with the gripper attached and no external payload, the robot achieved a maximum ascent speed of 56.2~mm/s and a maximum descent speed of 95.5~mm/s. The corresponding observed peak vertical accelerations in these trials were 362.3~mm/s$^2$ during ascent and 370.0~mm/s$^2$ during descent (Table~\ref{tab:performance_summary}).

\subsection{Attitude Control}
To quantify the attitude authority enabled by distributed buoyancy control, we measured the maximum statically stable tilt angle in water with the gripper module attached and no external payload. For each direction, we commanded the robot to increase its tilt until it could no longer maintain a stable equilibrium. The onboard IMU accelerometer readings $(a_x,a_y,a_z)$ were converted to a tilt angle \(\vartheta\) (angle between the body \(z\)-axis and gravity) using
\begin{equation}
\vartheta = \arccos\!\left(\frac{a_z}{\sqrt{a_x^2+a_y^2+a_z^2}}\right).
\end{equation}
We report the maximum statically stable tilt across single-engine and dual-engine ``engine-up'' configurations. Across the four single-engine cases (M1--M4, corresponding to $\pm Y$ and $\pm X$ directions), the maximum statically stable tilt reached $64.6^{\circ}$. Across the four dual-engine diagonal cases (M1+M2, M2+M3, M3+M4, M4+M1), the maximum statically stable tilt reached $61.8^{\circ}$. In a representative tilted-hover trial, the robot converged to an equilibrium tilt of 52.8$^\circ$ (0.921~rad); the peak tilt rate and tilt acceleration during the tilt transition were 0.332~rad/s and 0.528~rad/s$^2$, respectively (Table~\ref{tab:performance_summary}).

\begin{figure}[t]
  \centering
  \includegraphics[width=\columnwidth]{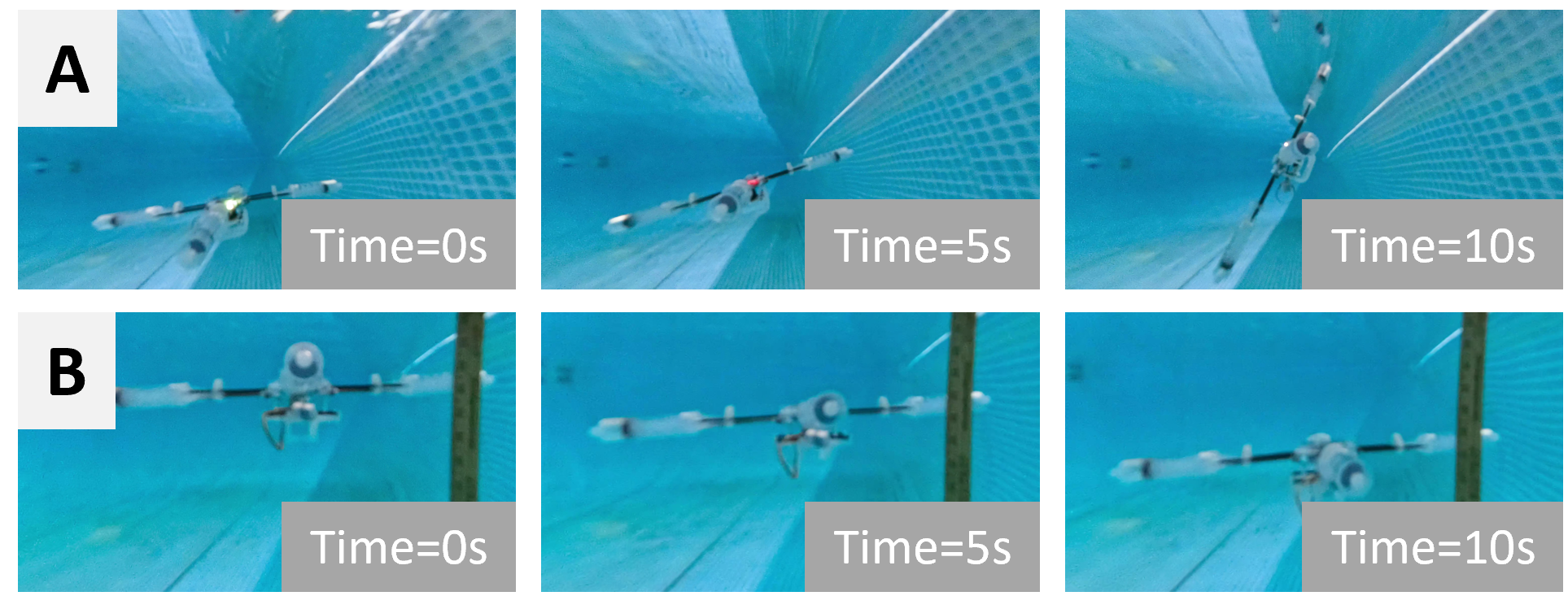}
  \caption{Representative locomotion primitives enabled by multi-engine buoyancy control. (A) Tilt regulation: the robot changes the relative displaced volume across engines to achieve and hold a nonzero tilt angle. (B) Descent regulation: the robot decreases net buoyancy (while maintaining attitude) to perform controlled sinking without thrusters.}
  \label{fig:locomotion}
\end{figure}

\begin{figure*}[t]
  \centering
  \includegraphics[width=\textwidth]{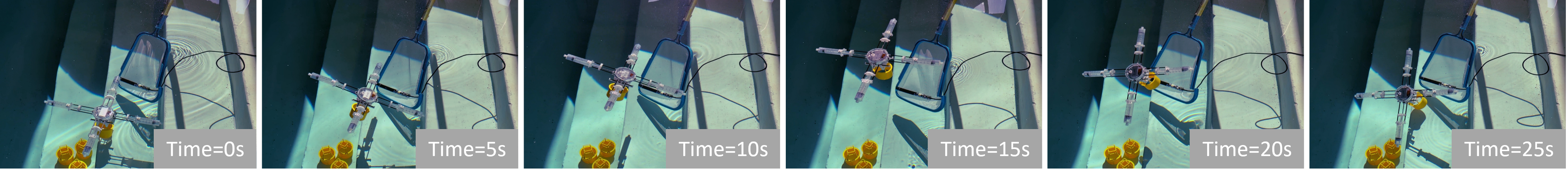}
  \caption{Thrusterless manipulation demonstration using buoyancy-driven tilt. A sequence of six snapshots (every 5~s) from \(t=0\)~s to \(t=25\)~s shows the gripper grasping a yellow payload and transporting it by alternating ascent and descent while regulating the vehicle tilt. The motion is achieved without thrusters by coordinating net buoyancy changes with attitude control.}
  \label{fig:manipulation}
\end{figure*}

\subsection{Gliding and Maneuvering}

Figure~\ref{fig:locomotion} shows two motion primitives used in our experiments: (A) tilt regulation through differential buoyancy commands and (B) controlled descent by decreasing net buoyancy. Combining these primitives with ascent enables thrusterless, glide-like motion; maintaining a nonzero tilt produced a visible horizontal component during ascent/descent (Fig.~\ref{fig:concept}(B)).

\subsection{Task Demonstrations}

Figure~\ref{fig:manipulation} demonstrates gripper-based payload transport. Starting from a grasped yellow payload at \(t=0\)~s, the robot alternates between tilted ascent and tilted descent to reposition the object in the water column while maintaining a stable grasp. This example illustrates how buoyancy-driven attitude and depth primitives can support payload transport without exposing propellers to entanglement risks.

\section{Limitations and Future Work}\label{sec:limitations}

While the platform demonstrates repeatable volume actuation and high-angle static buoyancy balance, several factors currently limit its performance. The high-level mapping from desired depth/tilt behaviors to per-engine setpoints is empirically tuned and may vary with payload properties, trapped air, water currents, and hydrodynamic interactions. Syringe-based mechanisms introduce seal friction, backlash, and hysteresis, which can limit actuation bandwidth and repeatability. Long-term operation also depends on sealing reliability and trapped-air management. Finally, because our experiments used offboard high-level commands and telemetry, underwater wireless communication remains a practical limitation; a waterproof antenna was placed near the robot in shallow water to maintain the UART link.

Despite these factors, the platform's modular magnetic interface and distributed buoyancy architecture offer a versatile foundation for diverse underwater applications. Beyond the demonstrated gripper, the system supports the rapid integration of onboard instrumentation, including cameras, lights, water-quality sensors, and task-specific end-effectors. A promising direction is \emph{in-water payload characterization}: because buoyancy is directly controlled via displaced volume, the robot can potentially estimate the buoyant weight of a grasped object by measuring the additional volume required to maintain neutral buoyancy or a fixed depth, enabling underwater ``weighing'' and in-situ property estimation. Another direction is to incorporate model-based or learning-based high-level control that adapts the setpoint mapping online across payloads and hydrodynamic conditions, supporting autonomous depth/tilt regulation and more complex behaviors. With reliable state estimation and closed-loop control, the platform can be extended to trajectory tracking and path planning for efficient thrusterless glide segments and manipulation sequences. The present prototype primarily regulates depth, pitch, and roll through distributed buoyancy and does not include a dedicated yaw actuator. For the demonstrated primitives, body-frame tilt direction was the primary controlled variable; yaw reorientation may be indirectly biased through cyclic roll--pitch transitions and hydrodynamic coupling, but precise heading regulation is left for future work. Communication can also be improved by adopting alternative links (e.g., tethered operation, optical links, or acoustic modems) and by designing protocols that tolerate packet loss and low update rates in underwater settings.

\section{Conclusion}
This paper presented \robotname, a multi-engine buoyancy-controlled underwater robot that enables thrusterless depth change, large whole-body tilting, and a magnetic modular payload interface. 
By modulating the center of buoyancy across four distributed engines, the robot achieves a total buoyancy authority of \(\approx\)160~g and maximum statically stable tilts exceeding 60$^\circ$.
Experimental characterization demonstrated peak vertical speeds of 95.5~mm/s and tilt rates of 0.332~rad/s, validating the robot's capacity for agile posture reconfiguration.
Furthermore, we demonstrated a gripper-based payload-transport sequence using these motion primitives, highlighting the potential of thrusterless operation for low-disturbance, entanglement-resistant underwater tasks.
Ultimately, this platform serves as a versatile foundation for future research in thrusterless underwater locomotion, autonomous multi-body manipulation, and in-situ environmental sensing in sensitive aquatic habitats.

\section*{Acknowledgment}
We used LLM tools, including Codex and GPT, to assist with software implementation, writing, proofreading, and literature search. All final text and references were edited and reviewed by humans.

\bibliographystyle{IEEEtran}
\bibliography{refs}

\end{document}